\documentclass[
	12pt, 
	a4paper,
]{scrartcl}

\newcommand{\graphic}{Graphics/}

\usepackage{geometry} 
\usepackage{scrlayer-scrpage}
\usepackage[T1]{fontenc}
\usepackage{lmodern} 

\usepackage{lscape}
\usepackage[format=plain,
labelfont={bf,it},
textfont=it]{caption}

\usepackage[british]{babel} 
\usepackage{textcomp}
\usepackage{enumitem} 
\usepackage{graphicx} 
\usepackage{eso-pic} 

\usepackage{tabularx} 
\usepackage{multirow} 
\usepackage{booktabs} 

\usepackage{amsmath,amssymb,amsfonts}
\usepackage{amsthm}
\usepackage{mathrsfs}
\usepackage{physics}

\usepackage{abstract}

\usepackage[hypertexnames=false]{hyperref} 
\usepackage{doi}
\usepackage[sort&compress, numbers, square]{natbib}

\usepackage{xcolor}
\usepackage{mdframed}

\usepackage{pdfpages}

\usepackage{float}

\usepackage{enumitem}

\definecolor{fraunhofergreen}{RGB}{23,156,125} 
\definecolor{fraunhoferblue}{RGB}{0,91,127} 
\definecolor{fraunhofergrey}{RGB}{166,187,200} 

\definecolor{fraunhoferorange}{RGB}{245,130,32} 

\definecolor{fraunhofergraphit}{RGB}{28,63,82} 
\definecolor{fraunhofersand}{RGB}{211,199,174} 
\definecolor{fraunhoferpetrol}{RGB}{0,133,152}
\definecolor{fraunhoferaqua}{RGB}{57,193,205}  
\definecolor{fraunhoferlime}{RGB}{178,210,53}
\definecolor{fraunhoferyellow}{RGB}{253,185,19}  
\definecolor{fraunhoferred}{RGB}{187,0,86}
\definecolor{fraunhoferwinered}{RGB}{124,21,77}

\newcolumntype{L}{>{\raggedright\arraybackslash}X}
\newcolumntype{R}{>{\raggedleft\arraybackslash}X}

\newcommand\entrynumberwithprefix[2]{#1\enspace#2:\hfill}

\DeclareTOCStyleEntry[
entrynumberformat=\entrynumberwithprefix{\figurename},
dynnumwidth,
numsep=1em
]{tocline}{figure}

\DeclareTOCStyleEntry[
entrynumberformat=\entrynumberwithprefix{\tablename},
dynnumwidth,
numsep=1em
]{tocline}{table}

\DeclareTOCStyleEntry[
beforeskip = 0.25cm
]{tocline}{subsection}

\newcommand{\infoboxsectiontitle}{Infoboxenverzeichnis} 

\DeclareNewTOC[
type=infobox,                         
types=infoboxes,                      
float,                                
floatpos=tbp,                         
name=Infobox,                         
listname={\infoboxsectiontitle},         
tocentrystyle=tocline,                
tocentrylevel=1,                      
tocentrynumwidth=23mm,                
tocentryentrynumberformat= \entrynumberwithprefix{\infoboxname} 
]{loi}

\addtokomafont{title}{\color{white}} 
\addtokomafont{sectioning}{\color{fraunhoferblue}} 
\addtokomafont{subsubsection}{\color{black}}
\addtokomafont{paragraph}{\color{black}}

\makeatletter
\renewcommand{\sectionlinesformat}[4]{	
	\Ifstr{#1}{section}{				
		\Huge							
		
		\@hangfrom{\hskip #2#3}{\parbox[t]{0.89\linewidth}{\raggedright#4 
				
				\rule[3mm]{15mm}{1.75mm}}	 
			
			\vspace{5mm}
		}%
	}{									
		\@hangfrom{\hskip #2#3}{#4\vspace{-5mm}}}	
}
\makeatother
\addtokomafont{subsubsection}{\small}

\addtokomafont{section}{\vspace{1.5cm}}

\setitemize{labelindent=1.5em,labelsep=0.5cm,leftmargin=*}

\clearpairofpagestyles 	

\ofoot{
	\raisebox{0pt}[\ht\strutbox][\dp\strutbox]{\color{fraunhofergreen}\rule[-8mm]{2pt}{15mm}}\;
	\pagemark
}

\setkomafont{pageheadfoot}{\normalfont\color{fraunhofergrey}} 
\automark[section]{section} 
\ohead{\headmark} 

\hypersetup{
	colorlinks,
	linkcolor={fraunhoferblue},
	citecolor={fraunhoferblue},
	urlcolor={fraunhoferblue}
}

\newcommand{\institutes}[1]{
	\def\institute{#1}
}

\makeatletter
\newcommand{\titlepagedesign}{
	\begin{titlepage}
		\begin{flushright}
			\hspace{-2.5cm}
			\includegraphics[scale=0.8]{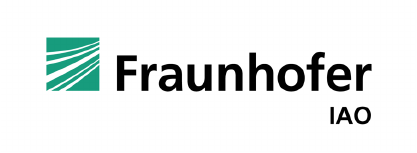}
		\end{flushright}
		\raggedright
		\AddToShipoutPictureBG*{
			\includegraphics[width=\paperwidth,height=\paperheight]{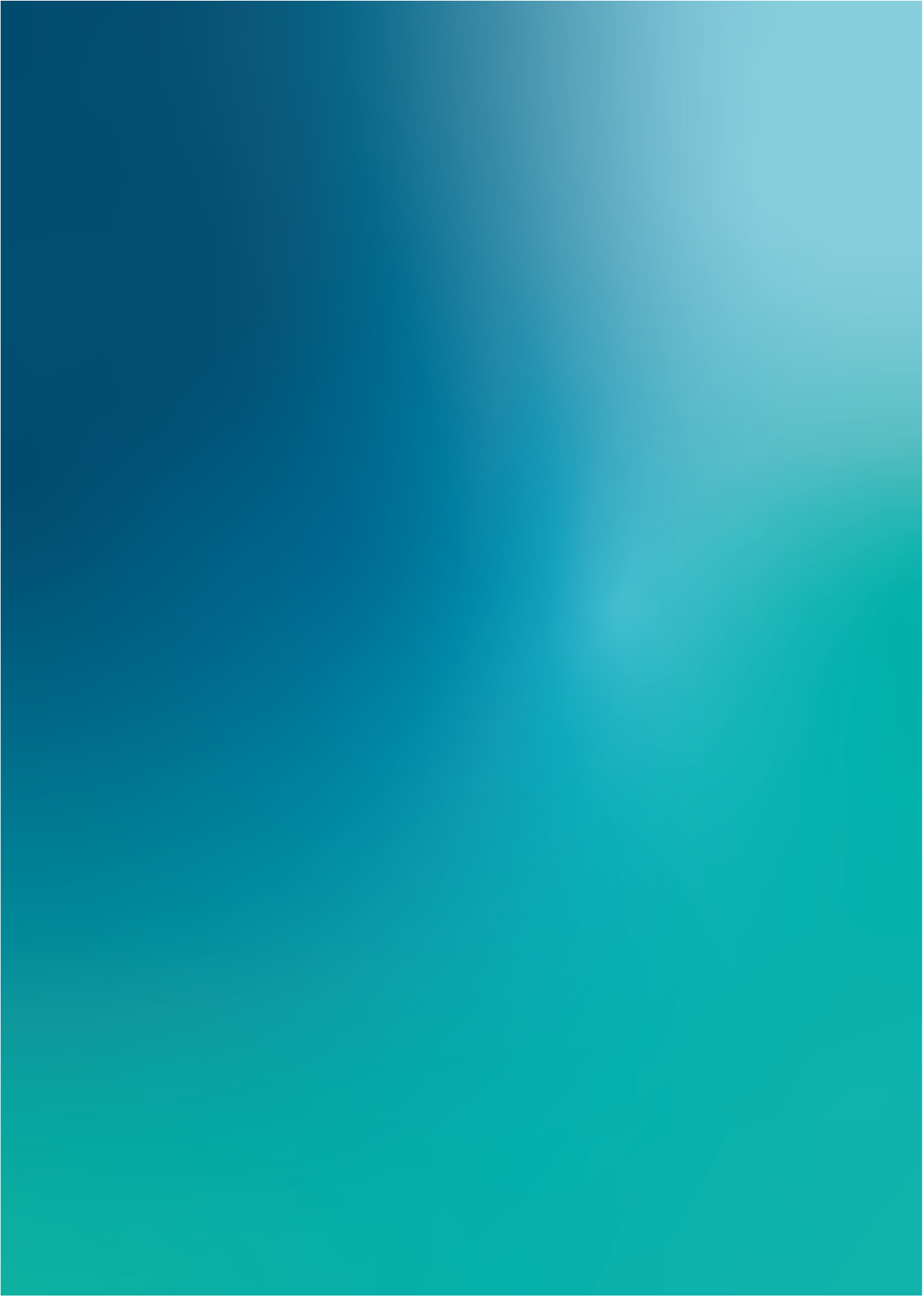}
		} 
		$~$
		
		\vfill

		\begin{center}
			\hspace{-1cm}
			\includegraphics[width = 0.5\paperwidth]{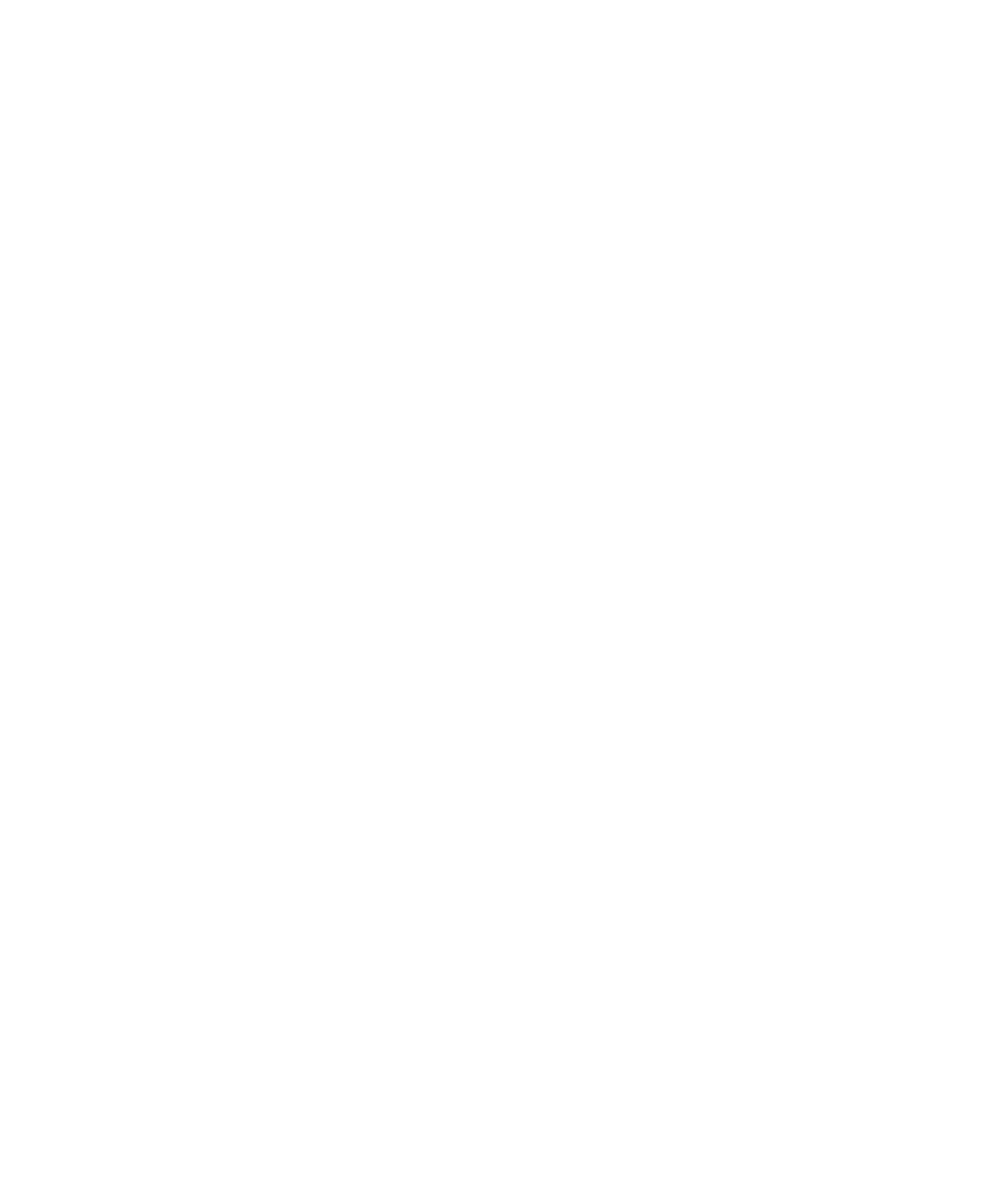} 
		\end{center}
		
		\vfill
		
		\color{white}
			\@author \\
			
			\institute
			
			\vspace{1mm}
			\Huge{\@title}\\
			\rule{11mm}{1.75mm}\\[2mm]
			\Large\textbf{{\@subtitle}}\\[6mm]
		
	\end{titlepage}
}
\makeatother

\author{%
	Dennis Klau\textsuperscript{1} | Marc Zöller\textsuperscript{2} | Dr. Christian Tutschku\textsuperscript{1}
}
\title{Bringing Quantum Algorithms to Automated Machine Learning}
\subtitle{%
	A Systematic Review of AutoML Frameworks\\
	Regarding Extensibility for QML Algorithms
}
\institutes{%
	\textsuperscript{1}\textmd{\textit{Fraunhofer IAO, Nobelstraße 12, 70569 Stuttgart, Germany}}\\ 
	\textsuperscript{2}\textmd{\textit{USU GmbH, Rüppurrer Str. 1, 76137 Karlsruhe, Germany}}
}
 
\usepackage{etoolbox}
\preto\section{%
	\ifnum\value{subsection}=0\addtocontents{toc}{\vskip10pt}\fi
}

\begin{document}

\pagestyle{empty} 

{%
	\newgeometry{left=2.5cm, right=1.5cm, top=1.5cm, bottom=2cm}
	\titlepagedesign
	\clearpage
	\restoregeometry
}

\begin{abstract}
    {\color{gray}
	Quantum Computing (QC) is becoming an increasingly promising technology for modern computation, especially in the field of data driven approaches like simulation and machine learning. With the high momentum of research and development of new hard- and software, QC holds a big promise in redefining many state-of-the-art computation approaches today.
	
	On the other hand, the nowadays well-established field of machine learning (ML) faces challenges like the discrepancy of demand by industry and availability of ML experts, reproducibility, and efficiency in prototyping. To overcome some of these issues, several frameworks have been created for automating the process of pipeline construction, data preprocessing, model training and hyperparameter optimization (HPO), many of them open source. In most cases, these Automated Machine Learning (AutoML) frameworks implement a fixed subset of known approaches and algorithms, or encapsulate an established ML backend, that defines the available algorithms.
	}
 
	This work describes the selection approach and analysis of existing AutoML frameworks regarding their capability of a) incorporating Quantum Machine Learning (QML) algorithms into this automated solving approach of the AutoML framing and b) solving a set of industrial use-cases with different ML problem types by benchmarking their most important characteristics. For that, available open-source tools are condensed into a market overview and suitable frameworks are systematically selected on a multi-phase, multi-criteria approach. This is done by considering software selection approaches \cite{ref1}, as well as in terms of the technical perspective of AutoML \cite{ref2, ref3}.
	
	The requirements for the framework selection are divided into hard and soft criteria regarding their software and ML attributes. Additionally, a classification of AutoML frameworks is made into high- and low-level types, inspired by the findings of \cite{ref4}. Finally, we select Ray and AutoGluon as the suitable low- and high-level frameworks respectively, as they fulfil all requirements sufficiently and received the best evaluation feedback during the use-case study.
	
	Based on those findings, we build an extended Automated Quantum Machine Learning (AutoQML) framework with QC-specific pipeline steps and decision characteristics for hardware and software constraints.
\end{abstract}
\newpage

\tableofcontents
\vspace{2cm}

{\normalcolor
\mdseries 
\subsection*{Editors}
Thomas Renner | Director Digital Business, Fraunhofer IAO\\
Prof. Dr.-Ing. Oliver Riedel | Executive Director Fraunhofer IAO
}
\newpage

\pagestyle{scrheadings} 

\setcounter{page}{1}
\section{Introduction \& Motivation}
\label{section1}

Building Machine Learning (ML) applications is a complex task, that usually requires involvement from both ML and domain experts. Additionally, the development of such systems is an iterative process, often driven by trial and error. So far, most of the required work to build ML applications is carried out by those experts manually: designing and implementing the individual steps from data cleaning, pre-processing and feature engineering, model tuning and hyper-parameter selection to evaluation and deployment. Consequently, building good pipelines can be a time and cost intensive task \cite{ref2}.

A similar situation is present in the Quantum Computing (QC) domain: Designing and tailoring quantum algorithms to a specific problem and testing different parametrizations on a simulated or real backend is usually done in an iterative manner by highly skilled experts, that require both, extensive domain and QC knowledge. Additionally, hardware handling is especially delicate still when computing on real backends: finding suitable quantum hardware for a specific problem and the associating data depends on multiple factors like current error rates and calibration, the coupling map, fidelity and decoherence times, as well as the speed of gates.

Automated Machine Learning (AutoML) is a potential solution to overcome some of the above-mentioned challenges that researchers, developers and adopters face \cite{ref4,ref3,ref5}. AutoML is an end-to-end approach for building ML applications, which aims to improve the iteration speed of and reduce the required knowledge for building ML pipelines via full or partial automation and encapsulation of algorithmic implementation details. This enables domain experts to create functioning first-draft ML pipelines, as well as increases efficiency of ML researchers and developers by automating tedious tasks like hyperparameter optimization (HPO) \cite{ref17}.

The general AutoML paradigm of knowledge and usability separation, as well as the increase of prototyping efficiency is also desired in the area of QC algorithm and software development, especially in the emerging subfield of Quantum Machine Learning (QML). This field follows the same concept of learning statistics from data as classical ML and is inspired by existing algorithms and implementations. For that, approaches from the ML domain are transferred to the QC paradigm, where equivalent, hybrid, or new versions of existing algorithms like kernel methods (e.g., QKE, QSVM \cite{ref6}) or neural networks (QNN, QC(C)NN \cite{ref7}) are developed.

To utilize the benefits of AutoML also in the QML domain, a review, classification and evaluation of existing AutoML frameworks regarding their capabilities and extensibility to quantum algorithms is mandatory. This includes the collection of the most prominent open-source frameworks to date with associated meta-data, the categorization of their degree of automation, optimization formulation and abstraction level. An evaluation of the most promising frameworks is done by solving specific use-cases and conducting expert interviews.

\newpage

The work is structured as follows: We start by giving an overview over the existing techniques and approaches in field of AutoML, followed by a classification of the existing frameworks, in Section~\ref{section2}. The motivation is to build a common understanding of this technology for the different target groups of AI and QC developers and researchers. In Section~\ref{section3}, we introduce the required extensions to the AutoML framing for the QC domain and introduce the terminology of AutoQML. This is followed by the framework selection approach and results – the core of our work – in Section~\ref{section4}. The selection approach is backed by the accompanied use-case study in Section~\ref{section5}, where identified frameworks are tested against multiple use-cases with different ML problem types. We conclude with a summary and future work perspective in Section~\ref{section6}.

\section{Systematic Review of AutoML}
\label{section2}
 
AutoML creates an abstraction layer between the goal formulation of an ML problem (e.g., image classification) with its associated data, and the knowledge level and programmatic details required to implement, train, and fine-tune a specific or range of different ML algorithms. 

\begin{figure}
    \includegraphics[width = \textwidth]{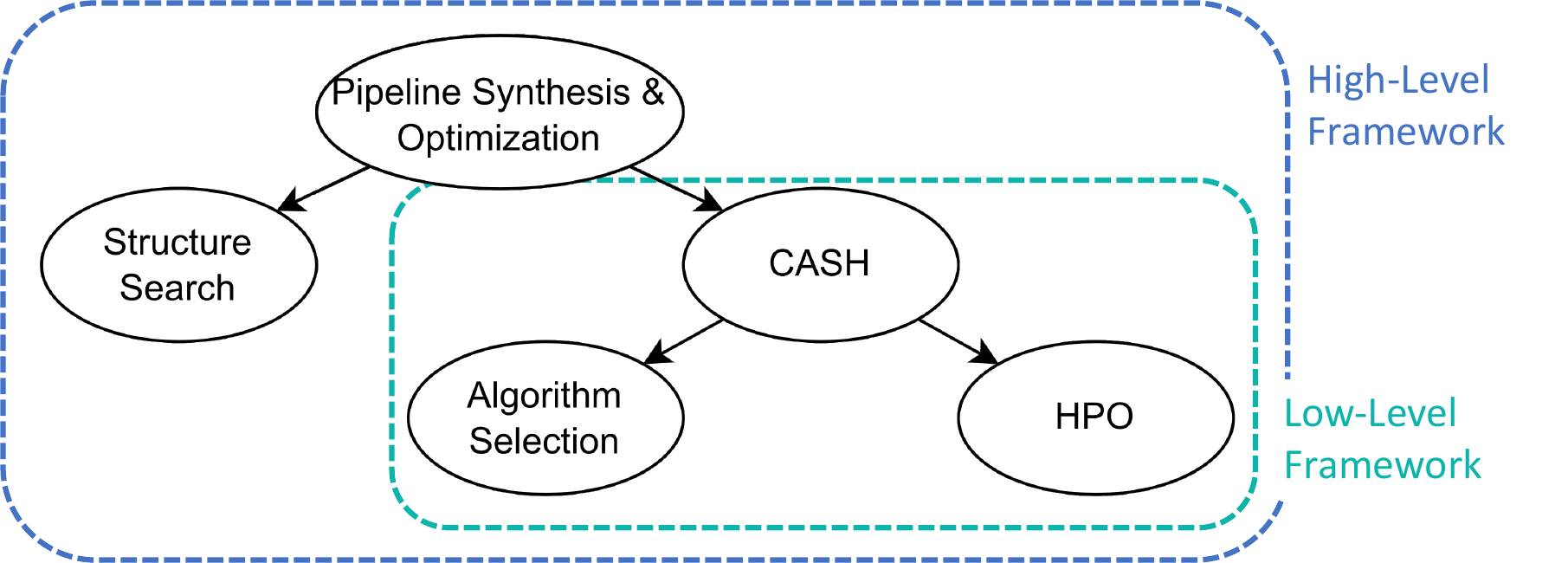}
    \caption{Stages and abstraction levels in the AutoML context. Parent nodes include the associated underlying automation and optimization tasks respective to the AutoML problem formulation. While the high-level frameworks (dark blue) try to optimize the whole ML pipeline, including its structure, appropriate pre-processing and cleaning steps for each algorithm and possible ensemble configurations, the low-level frameworks (green) assess mostly the CASH (combined algorithm selection and hyper-parameter optimization) or HPO (hyper-parameter optimization) paradigms. Taken from \cite{ref4}.}
    \label{fig1}
\end{figure}

\begin{figure}
	\includegraphics[width = \textwidth]{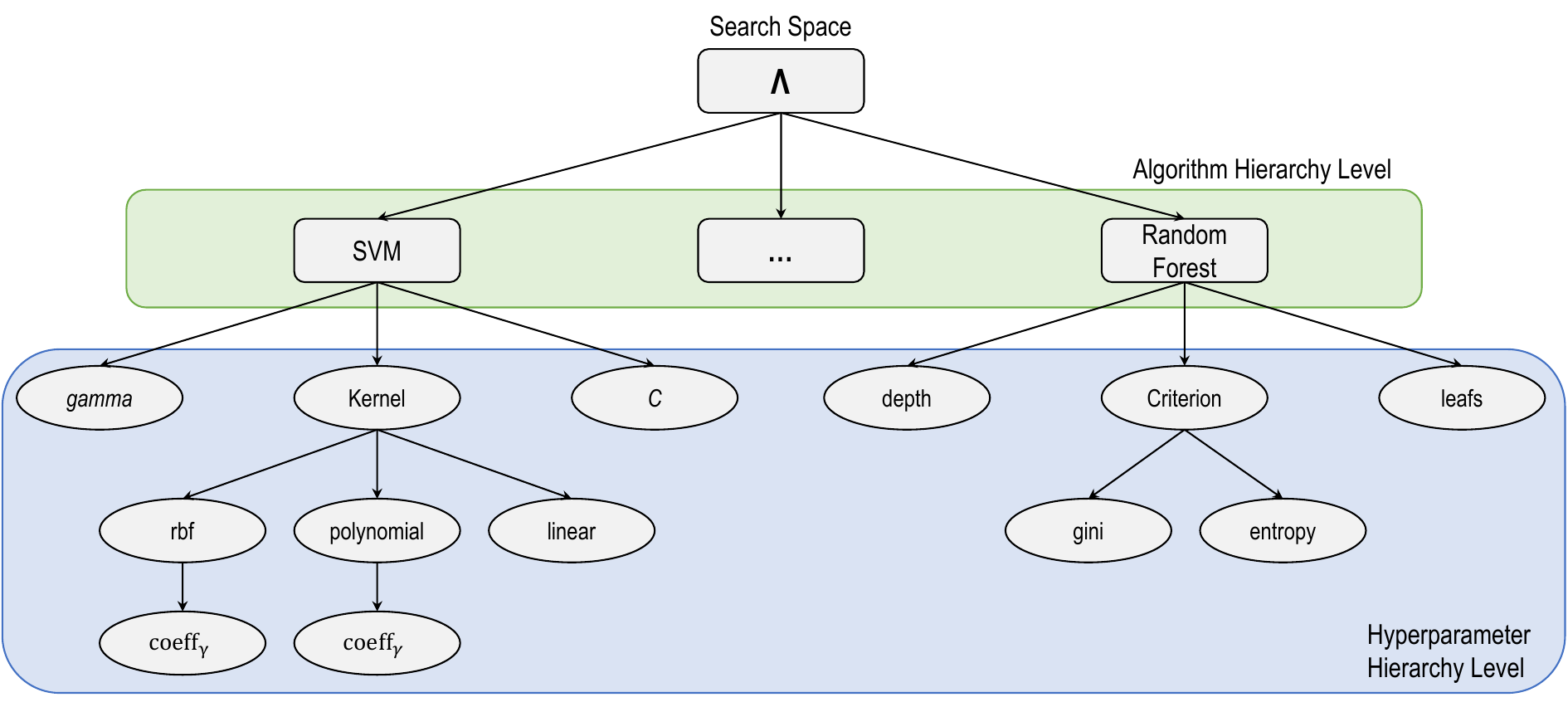}
	\caption{Tree-like hierarchical parameter structure for the Combined Algorithm Selection and Hyperparameter Optimization (CASH) formulation. From a defined global search space $\Lambda$ with all available hyperparameters, a hierarchical order must be constructed, that defines the subsets of meaningful hyperparameters to optimize during each iteration.}
	\label{fig2}
\end{figure}

Following the definition of \cite{ref4}, we classify the subproblems addressed by AutoML according to a tree-based structure as shown in Figure~\ref{fig1}. Going from a narrow to the broadest definition of tasks, the following subproblems can be partially or fully addressed by AutoML:

\begin{itemize}
\itemsep-2pt
	\item \textit{Hyperparameter Optimization (HPO)}: This is the smallest encapsulation of automation in the AutoML context. In HPO, the hyperparameters of a given algorithm are optimized \cite{ref2, ref8}. This can only be done outside of the actual training loop of the algorithm and requires a full or partial training cycle to assess the quality of the chosen hyperparameters. HPO is an computationally expensive task, thus in addition to naive approaches like grid search (GS) or random search (RS), also more sophisticated optimizers like Bayesian Optimization (BayesOpt) or evolutionary algorithms (e.g., DEHB \cite{ref9}) have been developed, that promise better convergence and less evaluations of the target function \cite{ref10}.
	
	\item \textit{Algorithm Selection (AS)}: Automatic AS is a fundamental aspect of pipeline construction, since no single algorithm defines the state-of-the-art for a given problem and usually a set of ML models with complementary strengths are tested. In the AutoML context with a given set of algorithms, AS is usually formulated as the per-instance algorithm selection problem \cite{ref11}.
	
	\item \textit{Combined Algorithm Selection and Hyperparameter Optimization (CASH}): is the natural extension of the optimization problem by combining AS and HPO. In the CASH setting, a set of available ML algorithms and their corresponding hyperparameters form a hierarchical search space – in the following denoted as $\Lambda$ - that can be traversed by an optimization algorithm, as shown in Figure~\ref{fig2}. An addition in CASH is the introduction of a search space structure in the sense that \textit{a priori} knowledge for HP combinations is available (i.e., that when optimizing the HP’s of a specific ML algorithm, the parameters of other algorithms are irrelevant and thus don’t need to be tested). This structure must be supported by the optimization algorithm, so that only valid and meaningful HP candidate sets are selected.
	
	\item \textit{Structure Search}: Since an ML pipeline normally involves many operations from data validation and cleaning, pre-processing, feature engineering and model execution, finding the optimal structure of a ML pipeline is a hard problem on its own. In addition to the pre-processing requirements of individual algorithms, also arbitrary ensemble models can be constructed which then lead to parallel pipeline paths. To find good ensemble configurations for specific problems, usually genetic algorithms or pre-defined pipeline structures are used.
	
	\item \textit{Pipeline Synthesis and Optimization (PSO)}: is the outer scope of the AutoML problem. It includes all previously described optimization approaches, i.e., performing Structure Search and CASH, of an ML pipeline as shown in Figure~\ref{fig1}.
\end{itemize}

To evaluate which frameworks address which scope of the AutoML cntext and are best suited for extension with QML algorithms, we additionally classify the considered frameworks with respect to their underlying abstraction and automation level into two types:

\begin{itemize}
	\item \textit{Low-level} frameworks require substantial expertise in data science and specifically machine learning. Such solutions need user-defined and (partially) structured search spaces, like algorithms to test and their respective hyperparameters, as well as already cleaned and preprocessed input data. These type of frameworks mostly build directly ontop of well-established ML frameworks (e.g., Scikit-Learn \cite{ref12}, PyTorch \cite{ref13}, Tensorflow \cite{ref14}) and wrap the basic algorithms inside of an automation routine with a scheduler and optimizer to efficiently apply approches from HPO up to CASH.
	\item \textit{High-level} frameworks, extend the approach of the low-level frameworks by attempting to solve the whole pipeline creation problem. This includes the low-level tasks (CASH or individual sub-tasks) as well as assembling a complete ML pipeline automatically. Since a ML pipeline consists of several dependent steps with high degrees of freedom, the search space has not only a predefined structure and is much larger but also more complex. The tasks for PSO usually include the inquiry of the architecture itself, selection of algorithms and their hyperparameters, as well as appropriate pre-processing and necessary data cleaning steps to some extend for each algorithm.
\end{itemize}

\begin{figure}
	\includegraphics[width = \linewidth]{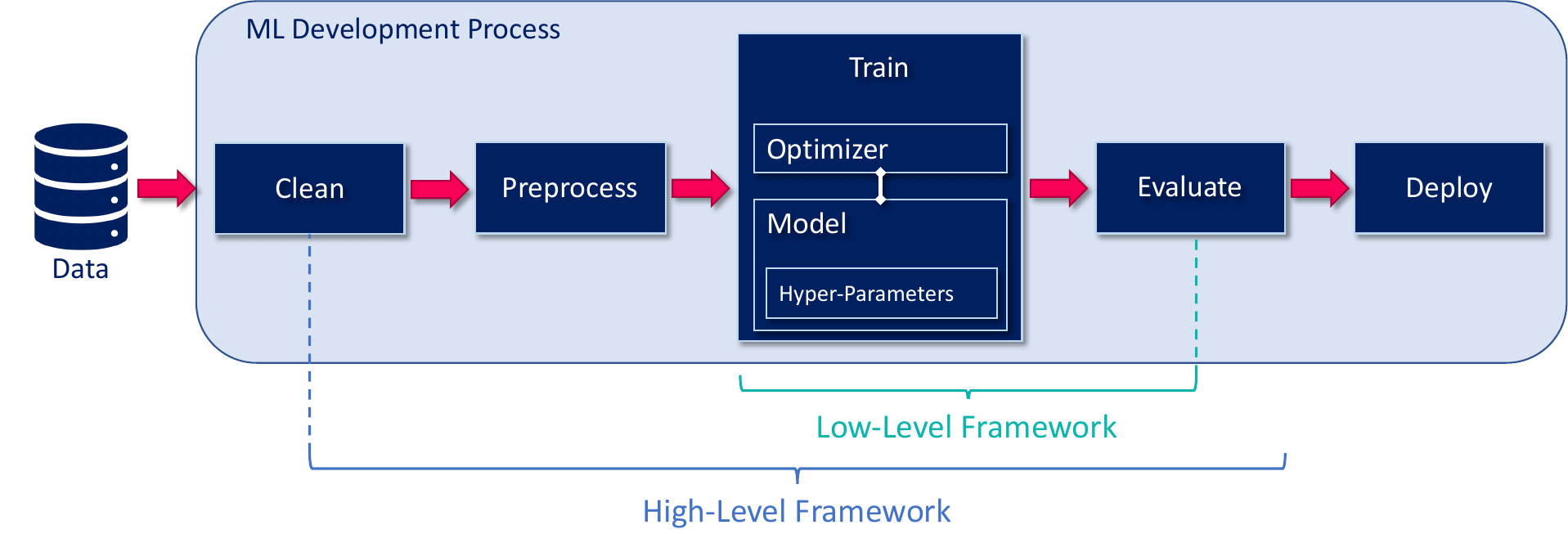}
	\caption{Depiction of a standard ML development process from data cleaning to model deployment. The different abstraction levels of AutoML address certain processing and evaluation steps in one pipeline. The low-level framework usually automates the classical training step by testing different combinations of algorithms and hyper-parameters. The high-level framework also encapsulates algorithm-dependent data pre-processing, as well as parts of the data cleaning procedure. Note that for simplicity, the construction of multiple vertically or horizontally stacked pipelines or pipeline parts (a technique that most high-level frameworks also utilize) is not depicted here.}
	\label{fig3}
\end{figure}

The approaches of each framework type are shown in Figure \ref{fig1}, as well as encapsulated steps in an exemplary ML pipeline in Figure~\ref{fig3}.

\section{Automated Quantum Machine Learning (AutoQML)}
\label{section3}

AutoML frameworks can be extended to incorporate quantum algorithms in multiple ways.
The synergy of integrating and utilizing QC within the AutoML context can be seen in multiple scenarios: (1) Using AutoML to select and configure suitable QML algorithms and (2) using quantum algorithms to improve the optimization process within AutoML. Finally, to execute the approaches above, (3) decision characteristics and meta-learning for automated QML can be implemented.
These approaches have different requirements inside of the AutoML modelling and thus require a diverse set of algorithms and pipeline elements, as well as frameworks that provide access those parts.


\begin{enumerate}[label=\textbf{\arabic*.}]
    \itemsep1em
	\item \textbf{Using AutoML methods to select and configure QML algorithms}\\
    For this approach, the user provides data to the AutoQML framework. This dataset could contain features from any domain, such as finance, health, or physics. As in the classical case, the framework is expected to analyse this dataset and draw conclusions about helpful preprocessing and feature engineering steps. Afterwards, QML algorithms are employed to train on the data. The framework is expected to select the most appropriate QML algorithm for the task and optimize its hyperparameters in an efficient and meaningful way. It then returns an optimized pipeline, which is a sequence of data processing elements, for the given problem.
    
    This approach describes the classical usage scenario of AutoML in the \textit{high-level} abstraction layer defined earlier. The user can solely provide the data and problem type for the task without the need to understand the intricate details of the underlying processing steps, applied QML algorithms, or the QC backends. This is highly advantageous for individuals who might not have expertise in ML or QC, as they can still utilize its capabilities without diving into the complexities.
    
    \item \textbf{Using quantum algorithms to improve AutoML optimization}\\
    In the second application, the focus shifts to using quantum computing to optimize the AutoML process itself. Since hyperparameters are crucial in determining the performance of an algorithm, AutoML frameworks usually search through a vast parameter search space $\Lambda$ (c.f. Figure \ref{fig2}) to find the best combination. Quantum computing has the potential to significantly expedite this search. By developing and testing quantum algorithms such as Quantum Bayesian Optimization \cite{ref15}, the AutoML framework can be enriched by alternative optimization approaches.
    
    This approach changes the optimizer used by the AutoML framework, which we normally do not have access to in the high-level frameworks. Thus, this use-case necessitates a more low-level interaction with the AutoML library. Here, the focus is on the optimization process itself and the user needs to understand (Q)ML algorithms, search spaces and hyperparameters. 
    
    \item \textbf{Developing decision characteristics and meta-learning for automated QML}\\
    Besides the integration of QC into the field of AutoML on the technical level, another important aspect of AutoQML is the essential aspect of determining how to make educated decisions on QML pipelines. This can be done by supporting developers in selecting QC algorithms and backends based on their characteristics, establishing meta-learning pipelines, tuning hyperparameters, and providing them with the tools to efficiently test and evaluate their quantum algorithms. The potential synergies of combining such automation approaches with the QC domain are:
    
    \begin{itemize}
    	\item QML algorithms have inherent attributes that make them more suitable for specific tasks like solving of linear systems, searching or combinatorial optimization due to the unique properties of quantum mechanics, such as superposition, entanglement, and quantum interference. Automatic decision criteria that recognize these properties can efficiently choose between classical and quantum algorithms for a given task.
    	\item Applying aspects of meta-learning like best-practice pipelines, where \textit{a priori} knowledge is either available or the system extracts it from previous tasks, the development of QML algorithms can benefit from past experiences. By studying prior QC runs and training sessions, the AutoML framework can either use refined and generalized rules or predict the best pipeline for new datasets, thus minimizing the computational cost and increasing efficiency.
    \end{itemize}
    
\end{enumerate}

Another important point is the growing need for tools that enable QC developers to test, evaluate, and benchmark their QML algorithms effectively and in a standardized way. This includes the ability to automatically run a set of (classical and QML) algorithms in their problem tailored configuration without much programming overhead, the comparison of performance metrics of those algorithms, and highlighting potential areas of improvement e.g. in terms of speed, accuracy, and resource consumption.


\section{Framework Selection Approach}
\label{section4}

\vspace{-8pt}
Software selection in general is a challenging process due to either the abundance or over-supply of options, limited time, and the complexity involved in understanding and evaluating these options. Integration with existing applications and meeting diverse requirements from predefined use-cases add to the complexity. Moreover, it's imperative to build consensus in tool handling and acceptance among the different user groups (AI and QC developers) as their differing expectations and requirements can influence the success of the chosen framework.

\begin{figure}[h]
	\includegraphics[width = \linewidth]{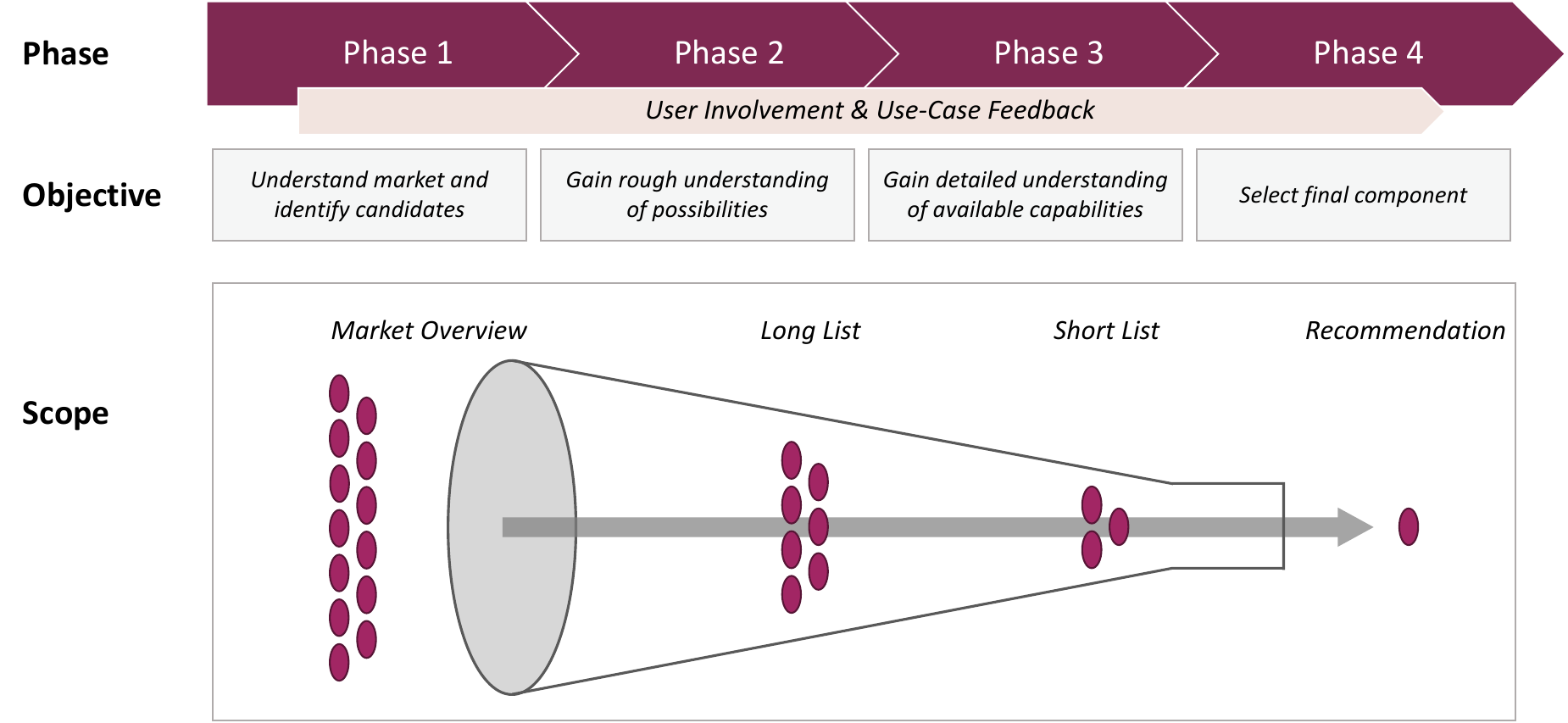}
	\caption{Framework selection approach following \cite{ref1}. The approach consists of four phases which gradually narrow down the available software tools to an appropriate recommendation. The different phases are described in detail in the text.}
	\label{fig4}
\end{figure}

For selecting the best-fitting framework, we draw inspiration from the software selection recommendations of Capgemini \cite{ref1} and use a four-phase approach, visualized in Figure~\ref{fig4}. The phases gradually increase in the number and detail of requirements to the evaluated frameworks and consist of the following steps:

\begin{itemize}
    \itemsep-7pt
	\item \textbf{Phase 1}: Following \cite{ref1}, we compile a \textit{“market overview”} of existing AutoML tools, which is a candidate list of available frameworks. Additionally, basic programmatic and technical requirements from both user sides are collected: the classical AI developers familiar with the ML algorithms and AutoML setting but lack in-depth knowledge in the quantum computing domain, and QC developers which usually have oppsite knowledge levels. These requirements are merged and added to the list as additional features. For each entry, we have classified the features into two categories: \textit{hard} and \textit{soft} features, which are distinct in the difficulty of acquisition and used to filter down the list in the following phases. This is done using a ranking system where each framework can collect points to reach a score between 1 (worst) and 5 (best). At the time of writing, no existing AutoML framework had native support for the special requirements of QC algorithms (described in phase 4), which will be implemented as wrappers around or inside the chosen frameworks.
	\item \textbf{Phase 2}: The overview of frameworks is roughly filtered in a first step by excluding those tools which do not meet the requirements of the hard feature set. Hard features are purely based on absolute metrics, that can be directly compared between the selected frameworks. Those features include the popularity (ranked by GitHub Stars), novelty and support (time of last commits and release, number of contributors), programming language and type of licence. Decision criteria for the hard features are: (1) must be under active development, (2) must have an active community, (3) must be open-source, (4) must have a Python interface, (5) must support the requirements of all considered real-world use-cases (described in Section 5), (6) must have a permissive license. Frameworks which do not meet the requirements are assigned a score of 1, the rest are added to a \textit{long list} of candidates.
	\item \textbf{Phase 3}: The \textit{long list} of frameworks is then further reduced using the collection soft features. Soft features are more diverse and include information only available after closer review of the documentation and / or source code. Those features are also not always directly comparable between the libraries, since e.g., different technology backends or algorithms are not necessarily better or worse than another solution. The collected features include implemented ML backends, supported input data types (e.g., tabular, images, etc.), supported ML problem types (e.g., classification, regression, etc.), type and size of the search space, implemented optimizers, and the categorization into the high- and low-level frameworks introduced in Section 2. Depending on how aligned they are to these requirements, frameworks are assigned a score between 2 – 5. Frameworks ranking the highest are added to the \textit{short list} of candidates.
	\item \textbf{Phase 4}: Frameworks included in the filtered short candidate list are in principle all a suitable choice, because they fulfill the collected hard and soft requirements. In addition to the prerequisites of the previous phases, two aspects are considered in the last phase: (1) feedback from the use-case studies regarding the experiences made with different applicable frameworks (see Section~\ref{section5}) and (2) implications of current framework limitations for QC. Since the objective is the integration of QC-native or -infused QML algorithms into the AutoML paradigm, specific challenges for quantum algorithms are collected from experts interviews and an estimation is made how well these challenges can be addressed by the frameworks included in the short candidate list. Current challenges for QC that shall be addressed and automated by AutoQML are mostly stemming from the fact, that QC today is still in a very early state, described as the noisy intermediate-scale quantum (NISQ) era. ML methods for today’s quantum hardware have to deal with a considerable amount of noise along with a limited number of qubits and short circuit depths. That necessitates feature reduction in pre-processing, post-processing in form of error mitigation, as well as extensive computation times and queuing on real QC backends. \\
    As decision criteria, we consider support and integration capabilities for QC-specific data encoding, backend selection and algorithm queuing as well as hardware constraints from the quantum computers themselves. Additionally, advanced analysis utility like automatic and meaningful feature selection (e.g., by dimensional reduction methods) are considered.
\end{itemize}

\subsection*{Selection Results}
\label{subsection4}

Our results of the phases 1 to 3 -- the collection of candidates as a market overview and the ranking of frameworks according to the hard and soft criteria -- are exemplary shown in Table~\ref{tab3} and the link to the full candidate list with the references to all considered frameworks is provided in Appendix~\ref{appendix1}. Additional considerations are made to later integrate the final framework into a platform ecosystem like PlanQK\footnote{PlanQK (Plattform und Ökosystem für Quantenapplikationen, \url{https://planqk.de}) is a German initiative for providing a platform for researchers and industry to exchange and provide quantum software solutions in an app-store like way.}, which is not specifically represented in the candidate list. 

Not all aspects of Section~\ref{section3} can be completely mapped to either a high- or low-level framework only. Thus, both framework types are required and the phases 3 and 4 are executed for both types respectively, resulting in a short list and final decision for both framework types. In our case, the short list after phase 3 consists of the following frameworks, which got a score of $\geq 4$:

\begin{table}[h]
\begin{center}
    \label{tab:selresult}
    \caption{Final \textit{short list} after the $3^{rd}$ phase for low- and high-level frameworks.}
	\begin{tabularx}{0.7\linewidth}{XX}
		\toprule
		\textbf{Low-level Frameworks} & \textbf{High-level Frameworks}\\
		\midrule[0.9pt]
		Ray & FLAML \\
		\midrule
		syne-tune & NNI\\
		\midrule
		Optuna & AutoGluon\\
		\midrule
		SMAC3 &\\
		\bottomrule
	\end{tabularx}
\end{center}
\end{table}

Afterwards, feedback from the participating use-case companies is collected regarding the required abstraction level of AutoML, usability, interpretability, framework constraints and performance of the tested candidate frameworks, described in detail in Section \ref{section5}. Additional consideration of the QC-specific requirements from phase 4 and feedback from the QC developer interviews described in Section \ref{subsection5.2} result in the final framework choices: 

\begin{itemize}
	\item \textbf{Ray} for the low-level framework and 
	\item \textbf{AutoGluon} for the high-level framework.
\end{itemize}

With an origin in parallel computation, Ray provides a highly flexible and performant solution for the CASH formulation with many optimizers to choose from in a plug-and-play style. Additionally, the framework is extremely popular, well maintained and documented. As a framework maintained and developed by Amazon, AutoGluon provides a similar level of support as well as state of the art PSO approaches like multi-level ensemble construction, which are missing in many other high-level frameworks.

For further development and extension of the chosen frameworks, we make use of the widely adopted design patterns of the Scikit-Learn library \cite{ref16}, used nowadays in many high-level libraries, adapters, and wrappers.

\section{Use-Case Study}
\label{section5}

In addition to the technical prerequisites and objectives for the framework selection described in Section~\ref{section4}, we conducted different use-case studies. The reason for that is two-fold: (1) collect real-world requirements from end-users for the hard feature set of phase 2 in the selection approach, and (2) validate the short list of candidates by implementing use-case specific proof-of-concepts (POC’s) using two or more of the candidate frameworks and give further recommendations for the final decision in phase 4. For that, each participating company was asked to implement a classical ML solution for their use-case using at least two of the AutoML frameworks identified in phase 3 in Section~\ref{subsection4}. Afterwards, an interview was conducted with the developers who implemented the POC’s using a standardized questionnaire. The reference to the full questionnaire is available in Appendix~\ref{appendix2}.

\subsection{Use-Case Description}
In total, four studies with companies from different backgrounds have been conducted\footnote{The use-case partners are also introduced here: \url{https://www.autoqml.ai/en/use-cases}}. Additional to the ML problem formulation and input data requirements, all use-case POC's are realised with one or multiple candidate frameworks. In the following, the partners and their respective use-case are described briefly, and their general ML requirements are listed in Table~\ref{tab2}.

\subsubsection{IAV GmbH Ingenieursgesellschaft Auto und Verkehr}
IAV is an engineering consultancy service provider in the mobility and IT sector. The use-case is the modelling and simulation of forces appearing at tires of cars, which from the ML perspective is a timeseries prediction task in the engine control unit. A digital twin shall be modeled that implements the behavior of the real unit. In a reduced problem formulation, there are 24 input and six output features. A time series prediction model, that operates as a filling prediction forecaster, is implemented.

\subsubsection{KEB Automation KG}
KEB is a system solver for automated drive technologies, also in the fields of mechanical and plant engineering. KEB Automation operates a transport system (AGILOX) for intralogistics tasks. Each AGILOX vehicle is a driverless vehicle that navigates through plants and warehouses to transport items. Occasionally, a vehicle may fail for operational or technical reasons, causing an immediate stop that requires human intervention and causes delays in operations. The AGILOX system generates data about the vehicle itself and order status and makes it available to other services. The goal is to develop an ML solution that classifies specific events on time series data.

\begin{table}[H]
	\centering
	\footnotesize
	\caption{Overview of the requirements and ML formulations of each use-case partner. Each partner implemented their use-case with multiple candidate frameworks, also resulting in the deduction of the required AutoML abstraction level (introduced in Section~\ref{section2}) for their specific use-case.}
	\label{tab2}
	\begin{tabularx}{\linewidth}{LLLLL}
		\toprule 
		& \textbf{IAV GmbH} & \textbf{KEB Automation KG} & \textbf{TRUMPF} & \textbf{Zeppelin GmbH} \\
		\midrule[0.9pt]
		\textbf{Prerequisites} & Own model definition must be possible & Commodity hardware & A CNN model shall be used & The solution should be easy to implement\\
		\midrule
		\textbf{Functional Requirements} & High quality documentation should be available & Real-time inference	 & Deep learning and accelerator hardware must be supported & Framework feedback and interpretability is important\\
		\midrule
		\textbf{Data Type(s)} & Time series (numerical) & Time series (numerical and categorical)  & Images (greyscale) & Tabular (numerical and categorical)\\
		\midrule
		\textbf{ML Problem} &
		Time Series Forecast&	Time Series Classification	&Image Classification	&Tabular Regression\\
		\midrule
		\textbf{Tested Frameworks}&
		Ray\newline
		Optuna	&
		Ray\newline
		Optuna\newline
		SMAC\newline
		Auto-Sklearn&
		Ray\newline
		AutoGluon	&
		Auto-Sklearn\newline
		FLAML\newline
		AutoGluon\newline
		Optuna\\
		\midrule
		\textbf{Resulting AutoML Abstraction Level} & Low-level & Low-level & Low-level & High-level\\
		\bottomrule
	\end{tabularx}
\end{table}

\subsubsection{TRUMPF Werkzeugmaschinen GmbH + Co. KG}
TRUMPF is a market and technology leader in machine tools and lasers, including their software solutions for industrial manufacturing as well as in the smart factory field. In the context of AutoQML, TRUMPF wants to optimize sheet metal processing by a laser cutting machine. Before the process of cutting, a metal sheet is first placed on the support bars of the pallet. The pallet is then moved into the machine room, where the cutting head is located and cuts the parts out of the sheet metal. The challenge is, that cut parts may tip over and cause a collision with the cutting head, interrupting the cutting process and potentially damaging the head. To avoid tilting, the cut parts should be stabilized by support bars. Those bars must be detected beforehand from a machine-attached camera using image data with convolutional neural networks (CNN) on the classical side.

\subsubsection{Zeppelin GmbH}
The Zeppelin Group offers solutions in the fields of construction, propulsion, and energy, as well as engineering and plant construction. Zeppelin Baumaschinen GmbH sales and services construction machinery. They also actively trade new and used machines. Trading with used machines requires the individual appraisal and evaluation of each machine that is to be bought or sold. This requires a high level of expertise, coupled with knowledge of current price developments, and usually leads to a valuation process lasting several hours. To support this process, an (Q)ML-powered price prediction tool is developed based on daily updated market data. This tool implements a regression model to predict the selling price of used construction machines. The importance of AutoML in this context was also shown previously in \cite{ref17}.

\subsection{Study Evaluation}\label{subsection5.2}
Here, we present some key messages and findings from the systematic interview evaluation. The complete study feedback (in German) is referenced in Appendix~\ref{appendix2}.

We evaluate the feedback from the POC's over all use-cases and with respect to their individual ML problem types stated in Table~\ref{tab2}. The answers to some questions can be subjective due to the fact of diverse prerequisites of each use-case, but also because of different familiarity levels in AutoML of the individual use-case experts and their preferences. Thus, the shown rankings are not directly generalizable, but serve as a first intuition and guideline for different applications with the same ML problem type. The prerequisites and functional requirements for each use-case is also provided to show different frame conditions. This is also the reason, why the same framework can occur on both sides of the ranking spectrum (see e.g., the Ray framework in the predictive power plot of Figure~\ref{fig5}).
The frameworks NNI and syne-tune were not tested by the use-case partners. 
The reason for NNI is that it only works on neural network algorithm classes, which was not general enough for all use-cases. syne-tune is the underlying native optimizer for the high-level framework AutoGluon, which was tested extensively.

\vspace{-8pt}
\subsubsection{Quality Aspects}
\begin{figure}
	\includegraphics[width=1\textwidth]{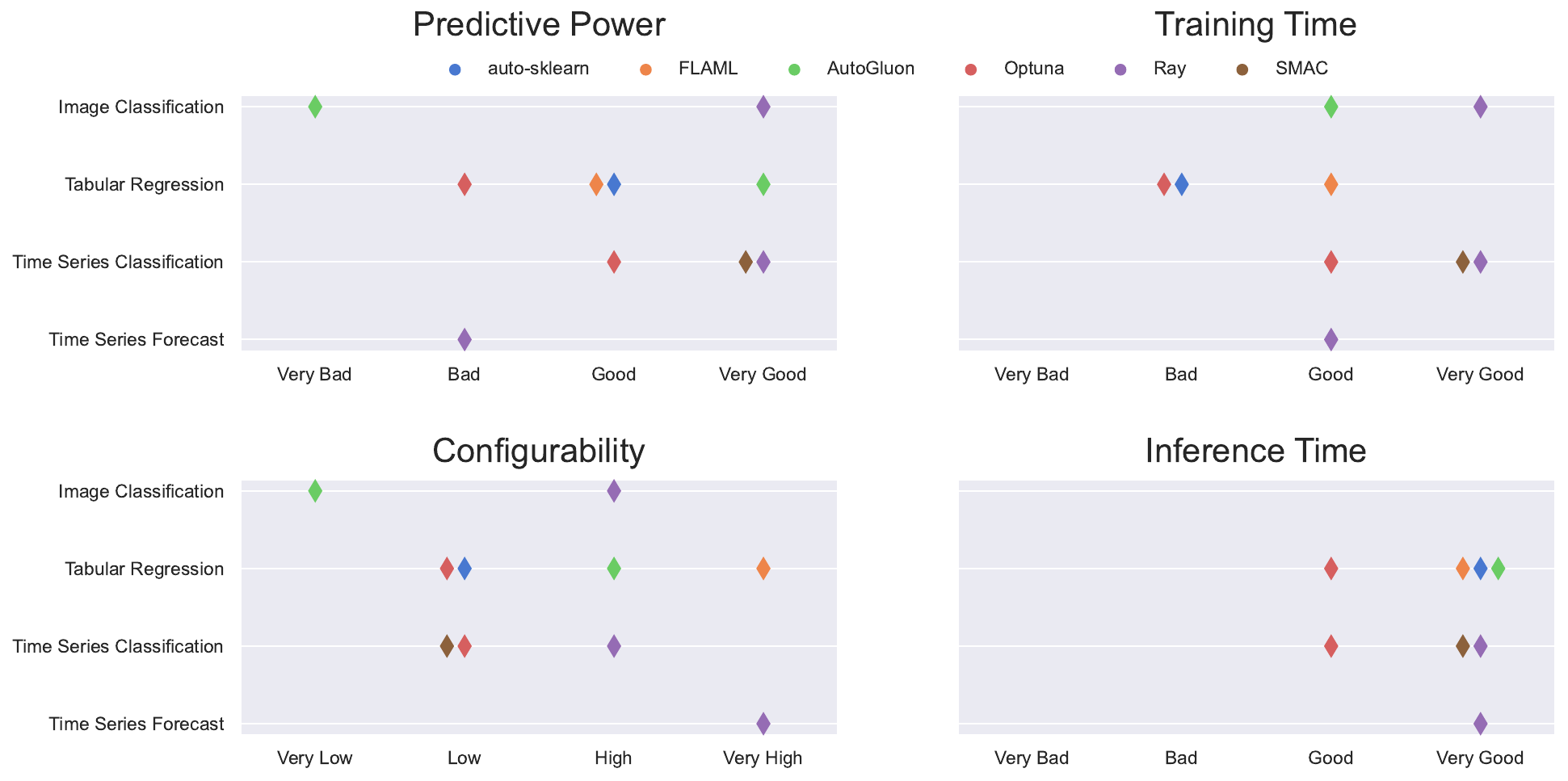}
	\caption{Evaluation results from the quality related assessments of the use-case study. Feedback is depicted for the final model performance (top left) and degree of configurability (bottom left), as well as the required training (top right) and inference time (bottom right) for each framework. Ratings are shown w.r.t. the respective ML problem type of each interview partner and their tested frameworks. If problem types have no or missing data points (e.g. inference time for the image classification task), the users reported that this has no relevance for their use-case and framework.}
	\label{fig5}
\end{figure}

In general, the predictive quality of the resulting models created by the tested frameworks was rated as good or very good by the ML experts in 70\% of all tested frameworks and use-case implementations, confirming the general potential of AutoML as a good approach for initial ML pipeline implementation. The high-level framework AutoGluon was reported to perform very bad in the image classification tasks in terms of resulting model performance and framework configuration possibilities, as shown in Figure~\ref{fig5}. This can mainly be attributed to the prerequisite of the use case: the experts perceived the task of incorporating and enforcing the use of specific models in the framework as difficult and partially not possible. On the other hand, the Ray framework was reported to achieve good overall results in every shown quality metric except for the model performance in the time series forecasting setting. Training times of the frameworks and inference speed of the resulting models were generally reported to be more than sufficient for the users among all use-cases. Only auto-sklearn and Optuna had bad training time requirements in the tabular regression setting. Our findings are in line with \cite{ref17}.

Especially the tested low-level frameworks, which only do HPO or CASH, comply with the predefined budget limit. PSO frameworks like auto-sklearn did not meet that requirement in some use-cases. In general, apart from sub-optimal implementation, we suspect this effect can occur due to the construction of whole pipelines which take longer to evaluate and makes violation of, e.g., time restrictions easier. From the evaluated use-case implementations, many partners reported that Optuna, while easy to use with a clean code basis, is too simple with too few SOTA optimizations for AutoML to deliver good results (both in configurability and model performance).

\subsubsection{Resource and API aspects}

\begin{figure}
	\includegraphics[width=\textwidth]{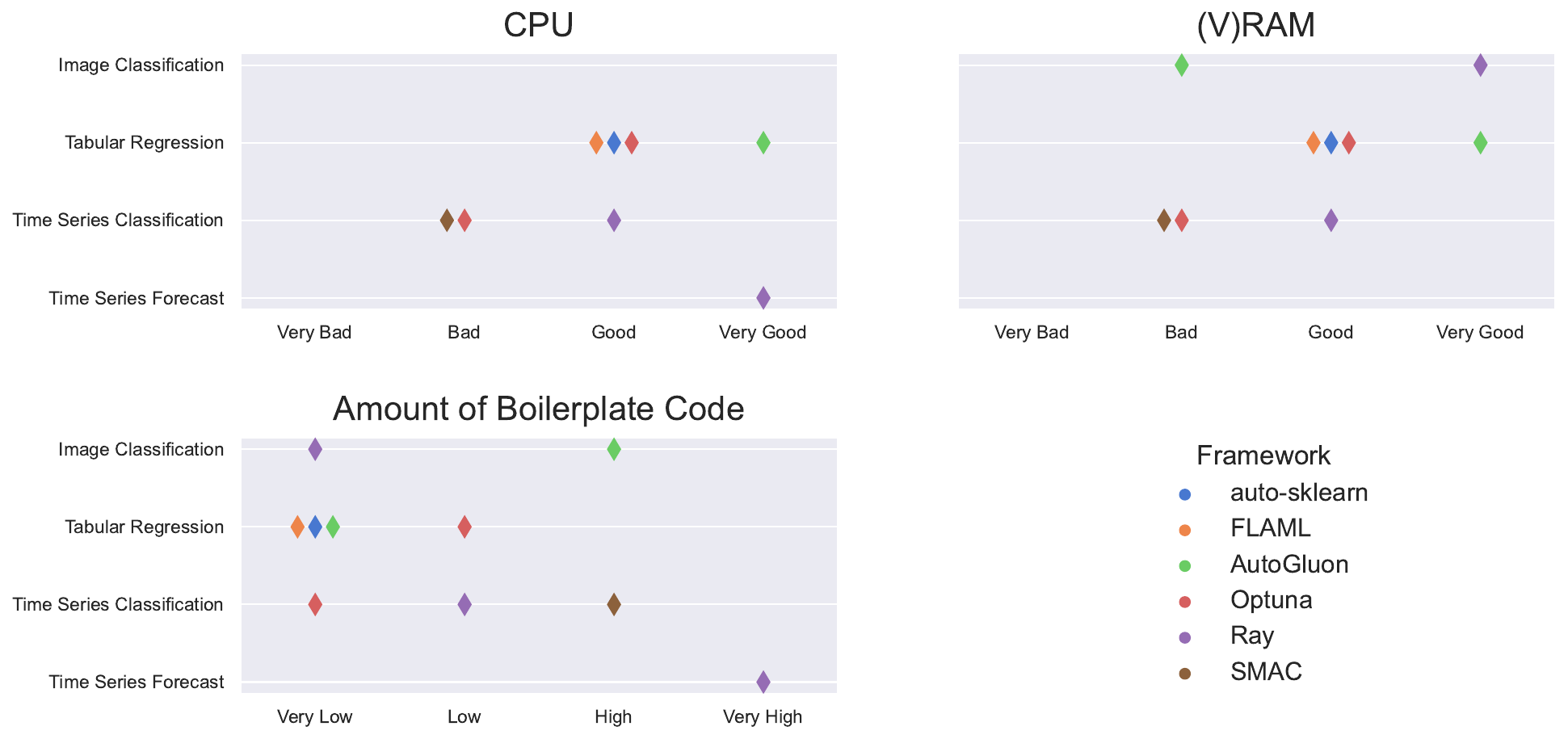}
	\caption{Interview feedback regarding resource and efficiency aspects of the different frameworks, classified by each ML use-case type. User ratings are shown in terms of CPU and (V)RAM utilization during training (top left and right). If a framework uses accelerator hardware, findings are reported for the combination of RAM and VRAM utilization. Additionally, the required amount of boilerplate / glue code is shown (bottom left), that was required by the developers to implement their use-case solution for each framework and the given requirements from Table \ref{tab2}. Not all use-cases gave feedback on every aspect or framework. A missing data point in the plot means that this aspect was not relevant for the implementation of the respective use-case.}
	\label{fig6}
\end{figure}

Most important computing resource demands (CPU, RAM / VRAM) are reported in Figure~\ref{fig6} (top left and right respectively). Resource consumption for AutoML frameworks is in general difficult to compare between different frameworks. Due to their mostly parallelized evaluation of many candidate configurations, they tend to be resource intensive programs with high demands for hardware, especially when optimizing deep learning models that utilize accelerator hardware. This was, however, expected or known by all developers. While mostly SMAC and Optuna were perceived to have unjustified high CPU and (V)RAM requirements, especially well established or recent frameworks have very good resource utilizations. The aspects of model size or complexity were mostly not relevant for the experts.

In general, the documentation, code examples and community help have been reported having high quality. The API of the frameworks was perceived easy to use, with less boilerplate or glue code necessary to create a working solution for the use-cases, except for Ray in some respect. This is shown in Figure~\ref{fig6} (bottom left). Particularly, low-level frameworks that require much more user interaction and adaption, or frameworks that did not fit to the use-case prerequisites, were reported to be more difficult to use. 

\newpage

\subsubsection{Quantum Aspects}
An individual interview was conducted with quantum computing and software development experts from Fraunhofer\footnote{Interviews were conducted with experts from the Fraunhofer instituts for "Industrial Engineering (IAO)" and "Manufacturing Engineering and Automation (IPA)"} regarding their future requirements when implementing quantum algorithms and utility functions into a chosen framework. A specification is, for example, that QML algorithms must be able to be included via wrapper functions into the existing framework, which also support HP configuration. Additionally, simulators must be supported as well as real QC hardware backends. Providing utility methods to automatically handle long queueing times and session handling is also required. Additionally, it should be possible to include external tool for evaluating the economical, execution and hardware restrictions of the submitted code. QC libraries that should be supported include IBM Qiskit\footnote{\url{https://www.ibm.com/quantum/qiskit-runtime}} and Xanadu’s PennyLane\footnote{\url{https://pennylane.ai}}, together with standard Python data science and AI packages.

These requirements mainly address the extensibility and code quality aspects of the reviewed frameworks and were additionally considered in the final choice as explained in Section \ref{subsection4}.

\vspace{2em}

\subsubsection{Additional Findings}
In general, the usability of all frameworks is high. While almost all the candidates in the short list are well documented, they come with some initial setup costs. The libraries usually have a high number of external dependencies making an integration into existing software environments more difficult. This holds especially for the high-level frameworks since they often include multiple backends and additional utility libraries. However, most of the use-case partners reported, that integration was quite easy due to the possibility to install the framework via package and dependency managers like \textit{pip} or \textit{conda}, which takes care of most of the package and version conflicts automatically.

The study also confirmed the automation motive of AutoML of \cite{ref17} in terms of required knowledge levels for different ML and data science tasks necessary to manually implement a ML pipeline: High-level frameworks like AutoGluon, FLAML and Auto-Sklearn don’t require the users to be proficient in the used ML algorithms or optimization strategies. The low-level frameworks like Ray, SMAC, and Optuna instead demand a more profound understanding of the individual algorithms, according to the interview feedback. The required knowledge for data preprocessing was reported high or very high for all tested frameworks, emphasizing the fact that every use-case had to implement additional preprocessing techniques and could not rely exclusively on the already available functionality, even in the high-level libraries. The necessary basic programming skills for each framework were reported with a high variance, leading to our conclusion that this is a highly subjective topic depending on the individual skills of the interview partner.

\newpage

\begin{landscape}
	$~$
	
\vfill
	\begin{table}[h]
		\caption{Excerpt of the framework candidate list. Frameworks, that didn’t meet the hard feature requirements got a rating value of 1 and were not analysed further. For example, Hyperband and Advisor were considered not under active development and community participation anymore due to their years of inactivity. H2O is written in Java and thus doesn’t provide the desired technology stack. The remaining frameworks were analysed in more detail and rated according to their suitability for the considered use-cases and extensibility.}
		\label{tab3}
		\tiny
		\begin{tabularx}{\linewidth}{lp{10mm}LLLllLp{11mm}p{23mm}p{19mm}p{22mm}Lr}
			\toprule
			\textbf{Framework} &\textbf{GitHub\newline Stars} & \textbf{Last Commit} & \textbf{Last Release} & \textbf{Nr. of Contributors} & \textbf{Language} & \textbf{License} & \textbf{Supp. ML Libraries} & \textbf{Input\newline Types} & \textbf{ML Problems} & \textbf{Search Space} & \textbf{Optimizer} & \textbf{Abstract. Level} & \textbf{Rating}\\
			\midrule[0.9pt]
			Hyperband & 572 & 13.09.2017 & & 1 & Python & BDS 2 & - & - & - & - & - & - & 1\\
			\midrule
			Advisor & 1500 & 11.11.2019 & 18.10.2018 & 9 & Python & Apache 2.0 & - & - & - & - & - & - & 1\\
			\midrule
			H2O 3 & 5900 & 04.08.2022 & 03.08.2022 & 160 & Java & Apache 2.0 & - & - & - & - & - & - & 1\\
			\midrule
			Auto-PyTorch & 1700 & 21.07.2022 & 18.07.2022 & 15 & Python & Apache 2.0 & PyTorch & Tabular\newline Data\newline Time\newline Series & Classification\newline Regression\newline Time Series Forecast & Neural Networks & Random Search\newline BayesOpt\newline Portfolios & High & 2\\
			\midrule
			TPOT & 8700 & 29.07.2022 & 06.01.2021 & 73 & Python & LGPL 3.0 & sklearn & Tabular\newline Data & Classification \newline Regression & Preprocessing\newline Classifier\newline Regressor & Evolutionary & High / Low & 2\\
			\midrule
			Auto-Sklearn & 6400 & 03.08.2022 & 18.02.2022 & 77 & Python & BSD 3 & sklearn & Tabular\newline Data & Classification\newline Regression & Preprocessing\newline Classifier\newline Regressor & smac & High & 2\\
			\midrule
			Hyperopt & 6300 & 29.11.2021 & 17.11.2021 & 90 & Python & Custom & Arbitrary & Arbitrary & Arbitrary & Arbitrary & BayesOpt & Low & 3\\
			\midrule
			Ray & 21400 & 04.08.2022 & 09.06.2022 & 709 & Python & Apache 2.0 & Arbitrary & Arbitrary & Arbitrary & Arbitrary & Grid Search\newline Random Search\newline Evolutionary\newline Other AutoML libs & Low & 5\\
			\midrule
			AutoGluon & 4700 & 03.08.2022 & 29.07.2022 & 81 & Python & Apache 2.0 & sklearn\newline PyTorch & Tabular\newline Data\newline Images\newline Text & Classification\newline Regression & Classifier\newline Regressor\newline Neural Networks & syne-tune\newline ray-tune & High & 5\\
			\midrule
			Optuna & 6700 & 03.08.2022 & 13.06.2022 & 175 & Python & MIT & Arbitrary & Arbitrary & Arbitrary & Arbitrary & Grid Search\newline Random Search\newline BayesOpt\newline Evolutionary & Low & 5\\
			\bottomrule
		\end{tabularx}
	\end{table}
	\vfill
	
	$~$
\end{landscape}
\section{Summary \& Future Work}
\label{section6}

In this work, we presented our systematic approach for the multi-criteria selection and assessment process of AutoML frameworks regarding their capability of solving real-world use-cases and extensibility with quantum computing algorithms. For that, we classified the most relevant AutoML frameworks into high- and low-level types. Following the software selection approach proposed by Capgemini \cite{ref1}, we adapted each phase to our specific needs. While collecting requirements from both end-user sides (ML and QC developers), we in parallel conducted a use-case study with four different use-cases and assessed the fitness of candidate frameworks by conducting expert interviews with the developers of each POC. The types of use-cases belong to the ML domains of classification, regression and forecasting, and work with tabular, time series and image input data. We evaluated the candidate frameworks during the selection phases on the use-cases based on quality aspects of the produced solution (e.g. predictive power, configurability, time requirements) and efficiency dimensions (e.g. hardware utilization, quality-of-life aspects). 

Our results showed, that there is a need for having functionalities from both: high- and low-level framework types. Especially the resulting performance quality was satisfactory for most use-cases and tested frameworks. After consideration of all requirements and the study feedback, the final choices for the AutoQML framework are Ray (low abstraction level) and AutoGluon (high abstraction level) due to their fulfillment of all requirements during the selection phases and overall best performance in the use-case evaluation. Supplementary material is provided on GitHub (see Appendix~\ref{appendix}).

Following this assessment, an open-source AutoQML framework will be implemented that combines AutoML and QC. For that, stand-alone QML algorithms will be developed which will be later integrated directly or as an independent backend into one of the existing AutoML libraries discussed above. Additionally, the chosen AutoML framework(s) will be extended by QC-specific preprocessing, search space and training requirements, which can either be integrated into the open-source software, or included as a wrapper library which utilizes the underlying automation functions.

\subsection*{Acknowledgements}

We thank the companies IAV, KEB, TRUMPF, Zeppelin and especially the developers for implementing the POC's and providing valuable feedback during the use-case study.

This work was created in the context of the AutoQML project, funded by the Federal Ministry for Economic Affairs and Climate Action. The project website can be found via this URL: \url{https://www.autoqml.ai/en}
\section{Appendix}
\label{appendix}

\subsection{Candidate list of open-source frameworks}
\label{appendix1}
The full candidate list is too long to be included in this document. Thus, the full Excel list of candidates, together with the hard and soft selection features and the framework scores can be accessed in this \href{https://github.com/AutoQML/Whitepaper-Framework-Selection/blob/main/doc/AutoML-Framework-Overview.xlsx}{GitHub repository}\footnote{\url{https://github.com/AutoQML/Whitepaper-Framework-Selection/blob/main/doc/AutoML-Framework-Overview.xlsx}}.

\subsection{Questionnaire of the ML expert interviews} 
\label{appendix2}
The excerpt of the MS Forms questionnaire used in the ML expert interviews with the use-case partners, together with the respective evaluation results, is available in the \href{https://github.com/AutoQML/Whitepaper-Framework-Selection/blob/main/doc/Interviews/Questionnair_Framework-Use-Case-Study.pdf}{GitHub repository}\footnote{\url{https://github.com/AutoQML/Whitepaper-Framework-Selection/blob/main/doc/Interviews/Questionnair_Framework-Use-Case-Study.pdf}}.


\phantomsection 
\addcontentsline{toc}{section}{\refname} 
\nocite{apsrev41Control}
\bibliography{masterfile, revtex-custom} 

\begin{thebibliography}{17}%
\makeatletter
\providecommand \@ifxundefined [1]{%
 \@ifx{#1\undefined}
}%
\providecommand \@ifnum [1]{%
 \ifnum #1\expandafter \@firstoftwo
 \else \expandafter \@secondoftwo
 \fi
}%
\providecommand \@ifx [1]{%
 \ifx #1\expandafter \@firstoftwo
 \else \expandafter \@secondoftwo
 \fi
}%
\providecommand \natexlab [1]{#1}%
\providecommand \enquote  [1]{``#1''}%
\providecommand \bibnamefont  [1]{#1}%
\providecommand \bibfnamefont [1]{#1}%
\providecommand \citenamefont [1]{#1}%
\providecommand \href@noop [0]{\@secondoftwo}%
\providecommand \href [0]{\begingroup \@sanitize@url \@href}%
\providecommand \@href[1]{\@@startlink{#1}\@@href}%
\providecommand \@@href[1]{\endgroup#1\@@endlink}%
\providecommand \@sanitize@url [0]{\catcode `\\12\catcode `\$12\catcode `\&12\catcode `\#12\catcode `\^12\catcode `\_12\catcode `\%12\relax}%
\providecommand \@@startlink[1]{}%
\providecommand \@@endlink[0]{}%
\providecommand \url  [0]{\begingroup\@sanitize@url \@url }%
\providecommand \@url [1]{\endgroup\@href {#1}{\urlprefix }}%
\providecommand \urlprefix  [0]{URL }%
\providecommand \Eprint [0]{\href }%
\providecommand \doibase [0]{http://dx.doi.org/}%
\providecommand \selectlanguage [0]{\@gobble}%
\providecommand \bibinfo  [0]{\@secondoftwo}%
\providecommand \bibfield  [0]{\@secondoftwo}%
\providecommand \translation [1]{[#1]}%
\providecommand \BibitemOpen [0]{}%
\providecommand \bibitemStop [0]{}%
\providecommand \bibitemNoStop [0]{.\EOS\space}%
\providecommand \EOS [0]{\spacefactor3000\relax}%
\providecommand \BibitemShut  [1]{\csname bibitem#1\endcsname}%
\let\auto@bib@innerbib\@empty
\bibitem [{\citenamefont {Middendorf}\ and\ \citenamefont {Kamann}(2016)}]{ref1}%
  \BibitemOpen
  \bibfield  {author} {\bibinfo {author} {\bibfnamefont {F.}~\bibnamefont {Middendorf}}\ and\ \bibinfo {author} {\bibfnamefont {G.}~\bibnamefont {Kamann}},\ }\href@noop {} {\emph {\bibinfo {title} {Software Selection: Managing the complexity of choosing the right software}}}\ (\bibinfo  {publisher} {Capgemini Consulting},\ \bibinfo {year} {2016})\BibitemShut {NoStop}%
\bibitem [{\citenamefont {Hutter}\ \emph {et~al.}(2019)\citenamefont {Hutter}, \citenamefont {Kotthoff},\ and\ \citenamefont {Vanschoren}}]{ref2}%
  \BibitemOpen
  \bibfield  {author} {\bibinfo {author} {\bibfnamefont {F.}~\bibnamefont {Hutter}}, \bibinfo {author} {\bibfnamefont {L.}~\bibnamefont {Kotthoff}}, \ and\ \bibinfo {author} {\bibfnamefont {J.}~\bibnamefont {Vanschoren}},\ }\href@noop {} {\emph {\bibinfo {title} {Automated Machine Learning - Methods, Systems, Challenges}}}\ (\bibinfo  {publisher} {Springer},\ \bibinfo {year} {2019})\BibitemShut {NoStop}%
\bibitem [{\citenamefont {Zöller}\ \emph {et~al.}(2021)\citenamefont {Zöller}, \citenamefont {Nguyen},\ and\ \citenamefont {Huber}}]{ref3}%
  \BibitemOpen
  \bibfield  {author} {\bibinfo {author} {\bibfnamefont {M.-A.}\ \bibnamefont {Zöller}}, \bibinfo {author} {\bibfnamefont {T.-D.}\ \bibnamefont {Nguyen}}, \ and\ \bibinfo {author} {\bibfnamefont {M.~F.}\ \bibnamefont {Huber}},\ }\bibfield  {title} {\enquote {\bibinfo {title} {Incremental search space construction for machine learning pipeline synthesis},}\ }in\ \href@noop {} {\emph {\bibinfo {booktitle} {International Symposium on Intelligent Data Analysis}}}\ (\bibinfo  {publisher} {arXiv:2101.10951},\ \bibinfo {year} {2021})\BibitemShut {NoStop}%
\bibitem [{\citenamefont {Zöller}\ and\ \citenamefont {Huber}(2021)}]{ref4}%
  \BibitemOpen
  \bibfield  {author} {\bibinfo {author} {\bibfnamefont {M.-A.}\ \bibnamefont {Zöller}}\ and\ \bibinfo {author} {\bibfnamefont {M.~F.}\ \bibnamefont {Huber}},\ }\bibfield  {title} {\enquote {\bibinfo {title} {Benchmark and survey of automated machine learning frameworks},}\ }\href@noop {} {\bibfield  {journal} {\bibinfo  {journal} {Journal of artificial intelligence research}\ ,\ \bibinfo {pages} {409--472}} (\bibinfo {year} {2021})},\ \bibinfo {note} {arXiv:1904.12054}\BibitemShut {NoStop}%
\bibitem [{\citenamefont {Quanming}\ \emph {et~al.}(2018)\citenamefont {Quanming}, \citenamefont {Mengshuo}, \citenamefont {Yuqiang}, \citenamefont {Wenyuan}, \citenamefont {Yu-Feng}, \citenamefont {Wei-Wei}, \citenamefont {Qiang},\ and\ \citenamefont {Yang}}]{ref5}%
  \BibitemOpen
  \bibfield  {author} {\bibinfo {author} {\bibfnamefont {Y.}~\bibnamefont {Quanming}}, \bibinfo {author} {\bibfnamefont {W.}~\bibnamefont {Mengshuo}}, \bibinfo {author} {\bibfnamefont {C.}~\bibnamefont {Yuqiang}}, \bibinfo {author} {\bibfnamefont {D.}~\bibnamefont {Wenyuan}}, \bibinfo {author} {\bibfnamefont {L.}~\bibnamefont {Yu-Feng}}, \bibinfo {author} {\bibfnamefont {T.}~\bibnamefont {Wei-Wei}}, \bibinfo {author} {\bibfnamefont {Y.}~\bibnamefont {Qiang}}, \ and\ \bibinfo {author} {\bibfnamefont {Y.}~\bibnamefont {Yang}},\ }\bibfield  {title} {\enquote {\bibinfo {title} {Taking human out of learning applications: A survey on automated machine learning},}\ \ }(\bibinfo  {publisher} {arXiv:1810.13306},\ \bibinfo {year} {2018})\BibitemShut {NoStop}%
\bibitem [{\citenamefont {Stühler}\ \emph {et~al.}(2023)\citenamefont {Stühler}, \citenamefont {Zöller}, \citenamefont {Klau}, \citenamefont {Beiderwellen-Bedrikow},\ and\ \citenamefont {Tutschku}}]{ref17}%
  \BibitemOpen
  \bibfield  {author} {\bibinfo {author} {\bibfnamefont {H.}~\bibnamefont {Stühler}}, \bibinfo {author} {\bibfnamefont {M.~A.}\ \bibnamefont {Zöller}}, \bibinfo {author} {\bibfnamefont {D.}~\bibnamefont {Klau}}, \bibinfo {author} {\bibfnamefont {A.}~\bibnamefont {Beiderwellen-Bedrikow}}, \ and\ \bibinfo {author} {\bibfnamefont {C.}~\bibnamefont {Tutschku}},\ }\bibfield  {title} {\enquote {\bibinfo {title} {Benchmarking automated machine learning methods for price forecasting applications},}\ }in\ \href@noop {} {\emph {\bibinfo {booktitle} {Proceedings of the 12th International Conference on Data Science, Technology and Applications}}}\ (\bibinfo  {publisher} {Rome, Italy},\ \bibinfo {year} {2023})\BibitemShut {NoStop}%
\bibitem [{\citenamefont {Rebentrost}\ \emph {et~al.}(2014)\citenamefont {Rebentrost}, \citenamefont {Mohseni},\ and\ \citenamefont {Lloyd}}]{ref6}%
  \BibitemOpen
  \bibfield  {author} {\bibinfo {author} {\bibfnamefont {P.}~\bibnamefont {Rebentrost}}, \bibinfo {author} {\bibfnamefont {M.}~\bibnamefont {Mohseni}}, \ and\ \bibinfo {author} {\bibfnamefont {S.}~\bibnamefont {Lloyd}},\ }\bibfield  {title} {\enquote {\bibinfo {title} {Quantum support vector machine for big data classification},}\ }\href@noop {} {\bibfield  {journal} {\bibinfo  {journal} {Physical Review Letters}\ }\textbf {\bibinfo {volume} {113}},\ \bibinfo {pages} {5} (\bibinfo {year} {2014})}\BibitemShut {NoStop}%
\bibitem [{\citenamefont {Beer}\ \emph {et~al.}(2020)\citenamefont {Beer}, \citenamefont {Bondarenko}, \citenamefont {Farrelly}, \citenamefont {Osborne}, \citenamefont {Salzmann}, \citenamefont {Scheiermann},\ and\ \citenamefont {Wolf}}]{ref7}%
  \BibitemOpen
  \bibfield  {author} {\bibinfo {author} {\bibfnamefont {K.}~\bibnamefont {Beer}}, \bibinfo {author} {\bibfnamefont {D.}~\bibnamefont {Bondarenko}}, \bibinfo {author} {\bibfnamefont {T.}~\bibnamefont {Farrelly}}, \bibinfo {author} {\bibfnamefont {T.~J.}\ \bibnamefont {Osborne}}, \bibinfo {author} {\bibfnamefont {R.}~\bibnamefont {Salzmann}}, \bibinfo {author} {\bibfnamefont {D.}~\bibnamefont {Scheiermann}}, \ and\ \bibinfo {author} {\bibfnamefont {R.}~\bibnamefont {Wolf}},\ }\bibfield  {title} {\enquote {\bibinfo {title} {Training deep quantum neural networks},}\ }\href@noop {} {\bibfield  {journal} {\bibinfo  {journal} {Nature Communications}\ }\textbf {\bibinfo {volume} {11}},\ \bibinfo {pages} {808} (\bibinfo {year} {2020})}\BibitemShut {NoStop}%
\bibitem [{\citenamefont {Feurer}\ and\ \citenamefont {Hutter}(2019)}]{ref8}%
  \BibitemOpen
  \bibfield  {author} {\bibinfo {author} {\bibfnamefont {M.}~\bibnamefont {Feurer}}\ and\ \bibinfo {author} {\bibfnamefont {F.}~\bibnamefont {Hutter}},\ }\bibfield  {title} {\enquote {\bibinfo {title} {Hyperparameter optimization},}\ }in\ \href@noop {} {\emph {\bibinfo {booktitle} {Automatic Machine Learning: Methods, Systems, Challenges}}}\ (\bibinfo  {publisher} {Springer},\ \bibinfo {year} {2019})\ pp.\ \bibinfo {pages} {3--38}\BibitemShut {NoStop}%
\bibitem [{\citenamefont {Awad}\ \emph {et~al.}(2021)\citenamefont {Awad}, \citenamefont {Mallik},\ and\ \citenamefont {Hutter}}]{ref9}%
  \BibitemOpen
  \bibfield  {author} {\bibinfo {author} {\bibfnamefont {N.}~\bibnamefont {Awad}}, \bibinfo {author} {\bibfnamefont {N.}~\bibnamefont {Mallik}}, \ and\ \bibinfo {author} {\bibfnamefont {F.}~\bibnamefont {Hutter}},\ }\bibfield  {title} {\enquote {\bibinfo {title} {{DEHB}: Evolutionary hyperband for scalable, robust and efficient hyperparameter optimization},}\ }\href@noop {} {\  (\bibinfo {year} {2021})},\ \bibinfo {note} {arXiv:2105.09821}\BibitemShut {NoStop}%
\bibitem [{\citenamefont {Bergstra}\ \emph {et~al.}(2011)\citenamefont {Bergstra}, \citenamefont {Bardenet}, \citenamefont {Bengio},\ and\ \citenamefont {Kégl}}]{ref10}%
  \BibitemOpen
  \bibfield  {author} {\bibinfo {author} {\bibfnamefont {J.}~\bibnamefont {Bergstra}}, \bibinfo {author} {\bibfnamefont {R.}~\bibnamefont {Bardenet}}, \bibinfo {author} {\bibfnamefont {Y.}~\bibnamefont {Bengio}}, \ and\ \bibinfo {author} {\bibfnamefont {B.}~\bibnamefont {Kégl}},\ }\bibfield  {title} {\enquote {\bibinfo {title} {Algorithms for hyper-parameter optimization},}\ }\href@noop {} {\bibfield  {journal} {\bibinfo  {journal} {Advances in Neural Information Processing Systems}\ } (\bibinfo {year} {2011})}\BibitemShut {NoStop}%
\bibitem [{\citenamefont {Kerschke}\ \emph {et~al.}(2019)\citenamefont {Kerschke}, \citenamefont {Hoos}, \citenamefont {Neumann},\ and\ \citenamefont {Trautmann}}]{ref11}%
  \BibitemOpen
  \bibfield  {author} {\bibinfo {author} {\bibfnamefont {P.}~\bibnamefont {Kerschke}}, \bibinfo {author} {\bibfnamefont {H.}~\bibnamefont {Hoos}}, \bibinfo {author} {\bibfnamefont {F.}~\bibnamefont {Neumann}}, \ and\ \bibinfo {author} {\bibfnamefont {H.}~\bibnamefont {Trautmann}},\ }\bibfield  {title} {\enquote {\bibinfo {title} {Automated algorithm selection: Survey and perspectives},}\ }\href@noop {} {\bibfield  {journal} {\bibinfo  {journal} {Evolutionary computations}\ }\textbf {\bibinfo {volume} {27}},\ \bibinfo {pages} {3--45} (\bibinfo {year} {2019})}\BibitemShut {NoStop}%
\bibitem [{\citenamefont {Pedregosa}\ \emph {et~al.}(2011)\citenamefont {Pedregosa}, \citenamefont {Varoquaux}, \citenamefont {Gramfort}, \citenamefont {Michel}, \citenamefont {Thirion},\ and\ \citenamefont {Grisel}}]{ref12}%
  \BibitemOpen
  \bibfield  {author} {\bibinfo {author} {\bibfnamefont {F.}~\bibnamefont {Pedregosa}}, \bibinfo {author} {\bibfnamefont {G.}~\bibnamefont {Varoquaux}}, \bibinfo {author} {\bibfnamefont {A.}~\bibnamefont {Gramfort}}, \bibinfo {author} {\bibfnamefont {V.}~\bibnamefont {Michel}}, \bibinfo {author} {\bibfnamefont {B.}~\bibnamefont {Thirion}}, \ and\ \bibinfo {author} {\bibfnamefont {O.}~\bibnamefont {Grisel}},\ }\bibfield  {title} {\enquote {\bibinfo {title} {Scikit-learn: Machine learning in python},}\ }\href@noop {} {\bibfield  {journal} {\bibinfo  {journal} {Journal of Machine Learning Research}\ ,\ \bibinfo {pages} {2825--2830}} (\bibinfo {year} {2011})}\BibitemShut {NoStop}%
\bibitem [{\citenamefont {Paszke}\ \emph {et~al.}(2019)\citenamefont {Paszke}, \citenamefont {Gross}, \citenamefont {Massa}, \citenamefont {Lerer},\ and\ \citenamefont {Bradbury}}]{ref13}%
  \BibitemOpen
  \bibfield  {author} {\bibinfo {author} {\bibfnamefont {A.}~\bibnamefont {Paszke}}, \bibinfo {author} {\bibfnamefont {S.}~\bibnamefont {Gross}}, \bibinfo {author} {\bibfnamefont {F.}~\bibnamefont {Massa}}, \bibinfo {author} {\bibfnamefont {A.}~\bibnamefont {Lerer}}, \ and\ \bibinfo {author} {\bibfnamefont {J.}~\bibnamefont {Bradbury}},\ }\bibfield  {title} {\enquote {\bibinfo {title} {Pytorch: An imperative style, high-performance deep learning library},}\ }in\ \href@noop {} {\emph {\bibinfo {booktitle} {Advances in Neural Information Processing Systems 32}}}\ (\bibinfo  {publisher} {Curran Associates, Inc.},\ \bibinfo {year} {2019})\ pp.\ \bibinfo {pages} {8024--8035}\BibitemShut {NoStop}%
\bibitem [{\citenamefont {Abadi}\ \emph {et~al.}(2015)\citenamefont {Abadi}, \citenamefont {Agarwal}, \citenamefont {Barham}, \citenamefont {Brevdo}, \citenamefont {Chen}, \citenamefont {Citro},\ and\ \citenamefont {Corrado}}]{ref14}%
  \BibitemOpen
  \bibfield  {author} {\bibinfo {author} {\bibfnamefont {M.}~\bibnamefont {Abadi}}, \bibinfo {author} {\bibfnamefont {A.}~\bibnamefont {Agarwal}}, \bibinfo {author} {\bibfnamefont {P.}~\bibnamefont {Barham}}, \bibinfo {author} {\bibfnamefont {E.}~\bibnamefont {Brevdo}}, \bibinfo {author} {\bibfnamefont {Z.}~\bibnamefont {Chen}}, \bibinfo {author} {\bibfnamefont {C.}~\bibnamefont {Citro}}, \ and\ \bibinfo {author} {\bibfnamefont {G.~S.}\ \bibnamefont {Corrado}},\ }\enquote {\bibinfo {title} {Tensorflow: Large-scale machine learning on heterogeneous systems},}\ \ (\bibinfo {year} {2015})\ pp.\ \bibinfo {pages} {265--283}\BibitemShut {NoStop}%
\bibitem [{\citenamefont {Rapp}\ and\ \citenamefont {Roth}(2023)}]{ref15}%
  \BibitemOpen
  \bibfield  {author} {\bibinfo {author} {\bibfnamefont {F.}~\bibnamefont {Rapp}}\ and\ \bibinfo {author} {\bibfnamefont {M.}~\bibnamefont {Roth}},\ }\bibfield  {title} {\enquote {\bibinfo {title} {Quantum gaussian process regression for bayesian optimization},}\ }\href@noop {} {\  (\bibinfo {year} {2023})},\ \bibinfo {note} {arXiv:2304.12923}\BibitemShut {NoStop}%
\bibitem [{\citenamefont {Buitinck}\ \emph {et~al.}(2013)\citenamefont {Buitinck}, \citenamefont {Louppe}, \citenamefont {Blondel}, \citenamefont {Pedregosa}, \citenamefont {Mueller}, \citenamefont {Grisel}, \citenamefont {Niculae}, \citenamefont {Prettenhofer}, \citenamefont {Gramfort}, \citenamefont {Grobler}, \citenamefont {Layton}, \citenamefont {VanderPlas}, \citenamefont {Joly}, \citenamefont {Holt},\ and\ \citenamefont {Varoquaux}}]{ref16}%
  \BibitemOpen
  \bibfield  {author} {\bibinfo {author} {\bibfnamefont {L.}~\bibnamefont {Buitinck}}, \bibinfo {author} {\bibfnamefont {G.}~\bibnamefont {Louppe}}, \bibinfo {author} {\bibfnamefont {M.}~\bibnamefont {Blondel}}, \bibinfo {author} {\bibfnamefont {F.}~\bibnamefont {Pedregosa}}, \bibinfo {author} {\bibfnamefont {A.}~\bibnamefont {Mueller}}, \bibinfo {author} {\bibfnamefont {O.}~\bibnamefont {Grisel}}, \bibinfo {author} {\bibfnamefont {V.}~\bibnamefont {Niculae}}, \bibinfo {author} {\bibfnamefont {P.}~\bibnamefont {Prettenhofer}}, \bibinfo {author} {\bibfnamefont {A.}~\bibnamefont {Gramfort}}, \bibinfo {author} {\bibfnamefont {J.}~\bibnamefont {Grobler}}, \bibinfo {author} {\bibfnamefont {R.}~\bibnamefont {Layton}}, \bibinfo {author} {\bibfnamefont {J.}~\bibnamefont {VanderPlas}}, \bibinfo {author} {\bibfnamefont {A.}~\bibnamefont {Joly}}, \bibinfo {author} {\bibfnamefont {B.}~\bibnamefont {Holt}}, \ and\ \bibinfo {author} {\bibfnamefont {G.}~\bibnamefont {Varoquaux}},\ }\bibfield  {title} {\enquote {\bibinfo
  {title} {{API} design for machine learning software: experiences from the scikit-learn project},}\ }in\ \href@noop {} {\emph {\bibinfo {booktitle} {ECML PKDD Workshop: Languages for Data Mining and Machine Learning}}}\ (\bibinfo {year} {2013})\BibitemShut {NoStop}%
\end{thebibliography}%

\pagestyle{empty} 
\AddToShipoutPictureBG*{
	\includegraphics[width=\paperwidth,height=\paperheight]{\graphic bg.pdf}}
	
\phantomsection
\addcontentsline{toc}{section}{Contact} 
\color{white}
\mdseries 
\color{white}
\section*{\color{white} Contact}
\vspace{2cm}
\begin{minipage}{0.3\linewidth}
	\includegraphics[scale=0.7]{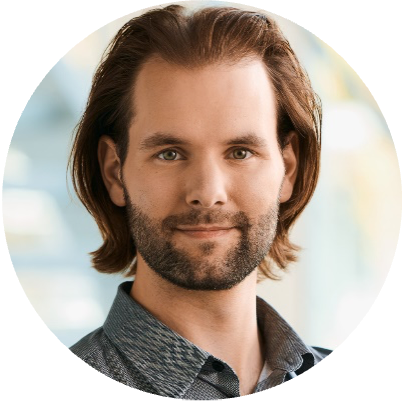}
\end{minipage}
\begin{minipage}{0.7\linewidth}
	\Large Dennis Klau\\
	Team Applied AI | Quantumcomputing\\
	dennis.klau@iao.fraunhofer.de\\
\end{minipage}	

\vspace{2cm}

\begin{minipage}{0.3\linewidth}
   \includegraphics[scale=0.7]{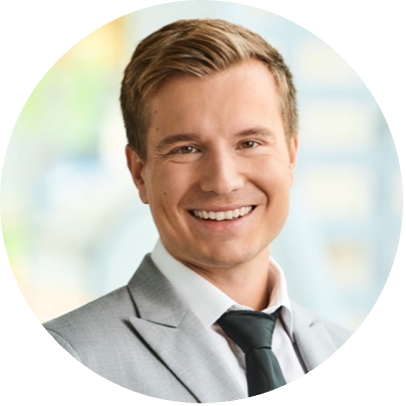}
\end{minipage}
\begin{minipage}{0.7\linewidth}
	\Large Dr. Christian Tutschku\\
	Teamlead Quantumcomputing\\
	christian.tutschku@iao.fraunhofer.de
\end{minipage}

\vspace{6cm}

\Large
Research Department Digital Business\\
Fraunhofer Institute for Industrial Engineering IAO\\
Nobelstraße 12, 70569 Stuttgart\\[10mm]
www.digital.iao.fraunhofer.de

\end{document}